\documentclass[fleqn,10pt]{wlscirep}

\usepackage[utf8]{inputenc} % allow utf-8 input
\usepackage[T1]{fontenc}    % use 8-bit T1 fonts
\usepackage{hyperref}       % hyperlinks
\usepackage{xr}
\usepackage{url}            % simple URL typesetting
\usepackage{booktabs}       % professional-quality tables
\usepackage{amsfonts}       % blackboard math symbols
\usepackage{nicefrac}       % compact symbols for 1/2, etc.
\usepackage{microtype}      % microtypography
\usepackage{xcolor}         % colors
\usepackage{graphicx}
\usepackage{amsmath}
\usepackage[utf8]{inputenc}
\usepackage[T1]{fontenc}
\usepackage{float}
\usepackage{booktabs}
\usepackage{colortbl}
\usepackage{titletoc}
\usepackage{xcolor}
\definecolor{darkgreen}{rgb}{0.0, 0.5, 0.0}
\definecolor{limegreen}{rgb}{0.5, 1.0, 0.0}
\definecolor{lightgreen}{rgb}{0.8, 1.0, 0.5}

\title{Beyond MD17: the reactive xxMD dataset}

\author[1]{Zihan Pengmei}
\author[2,3,4,5,6]{Junyu Liu}
\author[7,+]{Yinan Shu}

\affil[1]{Department of Chemistry, The University of Chicago, Chicago, IL 60637, USA}
\affil[2]{Pritzker School of Molecular Engineering, The University of Chicago, Chicago, IL 60637, USA}
\affil[3]{Department of Computer Science, The University of Chicago, Chicago, IL 60637, USA}
\affil[4]{Kadanoff Center for Theoretical Physics, The University of Chicago, Chicago, IL 60637, USA}
\affil[5]{qBraid Co., Chicago, IL 60615, USA}
\affil[6]{SeQure, Chicago, IL 60615, USA}
\affil[7]{Department of Chemistry, University of Minnesota, Minneapolis, MN 55414, USA}

\affil[+]{\url{shuxx055@umn.edu}}

\begin{abstract}
System specific neural force fields (NFFs) have gained popularity in computational chemistry. One of the most popular datasets as a bencharmk to develop NFFs models is the MD17 dataset and its subsequent extension. These datasets comprise geometries from the equilibrium region of the ground electronic state potential energy surface, sampled from direct adiabatic dynamics. However, many chemical reactions involve significant molecular geometrical deformations, for example, bond breaking. Therefore, MD17 is inadequate to represent a chemical reaction. To address this limitation in MD17, we introduce a new dataset, called Extended Excited-state Molecular Dynamics (xxMD) dataset. The xxMD dataset involves geometries sampled from direct non-adiabatic dynamics, and the energies are computed at both multireference wavefunction theory and density functional theory. We show that the xxMD dataset involves diverse geometries which represent chemical reactions. Assessment of NFF models on xxMD dataset reveals significantly higher predictive errors than those reported for MD17 and its variants. This work underscores the challenges faced in crafting a generalizable NFF model with extrapolation capability.
  
\end{abstract}
\begin{document}

\flushbottom
\maketitle
\thispagestyle{empty}
\section*{Background and Summary}
\subsection*{Introduction}
The development of molecular force fields driven by data is predominantly benchmarked against the MD17 dataset introduced by Chmiela et al. \cite{chmiela2017machine} and its extension, the rMD17 \cite{christensen2020role}. These datasets consist dynamic data of ten small to medium-sized gas-phase molecules. In molecular dynamics, data are intrinsically time-series sequences, necessitating careful sampling to prevent unintended information leakage into future states. A detailed analysis of MD17 and its variants reveals a significant sampling bias towards a narrow potential energy surface (PES) region close to the equilibrium structure. This narrow exploration of PES leads to limited conformation and energy space sampling, as our internal coordinate analysis shows. Thus, these datasets are suboptimal in terms of segmentation strategy and the molecular conformation space they cover.

For our discussion, we refer to these conventional molecular dynamics datasets as in-distribution (ID) datasets. Yet, many chemical processes of interest occur out-of-distribution. Consider a basic chemical reaction depicted in Figure \ref{fig:mpb}: the nuclear configuration space includes reactants, transition states, and products. Sampling exclusively from the reactant region fails to capture the full dynamics of chemical reactions. As a result, NFF models trained on such skewed datasets are biased towards reactant configurations, potentially leading to qualitatively inaccurate predictions for a complete chemical reaction.

To overcome these challenges, we introduce the extended excited-state molecular dynamics (xxMD) dataset in this work. The xxMD retains the core objective of capturing trajectory data for small to medium-sized gas-phase molecules but distinguishes itself by incorporating nonadiabatic trajectories which include the dynamics of excited electronic states. Comprising four photochemically active molecules, the xxMD begins with significantly higher initial energies, enabling it to traverse a more extensive nuclear configuration space and more authentically represent the entire chemical reaction PES — reactants, transition states, and products. Notably, the xxMD captures regions near conical intersections, which are critical to the pathways of potential energy surfaces across different electronic states.\cite{Tully1971,Truhlar1983, herman1984nonadiabatic,tully1990molecular,YarkonyCI,CIinMaterial} By including these key regions, the xxMD dataset aims to establish new benchmarks and challenges for NFF models, providing a more comprehensive and chemically accurate dataset for the development of predictive models.

We note that our development of xxMD datasets is not the first attempt ever to try to go beyond the (r)MD17 datasets. For example, the recently developed WS22 database \cite{w22} tries to include nuclear configurations from multiple minima and interpolate among these configurations. Although WS22 has gone beyond (r)MD17, the xxMD datasets developed in current work involve much more complex configurations, for example, regions that correspond to conical intersections and avoided crossings.

\subsection*{Existing datasets: MD17 and its variant}
Chmiela et al. performed adiabatic ab initio molecular dynamics (AIMD) simulations on small gas-phase molecules at room temperature, with the electronic potential energies computed at the Kohn-Sham density functional theory (KS-DFT) level\cite{chmiela2017machine}. However, the original publication did not provide detailed specifics about the density functional, basis set, spin-polarization, grid for integration, and the software used. This lack of transparency presents a challenge for reproducibility and may limit the utility of the dataset for certain types of chemical simulation.
\begin{figure}[ht]
\centering
\includegraphics[width=0.3\linewidth]{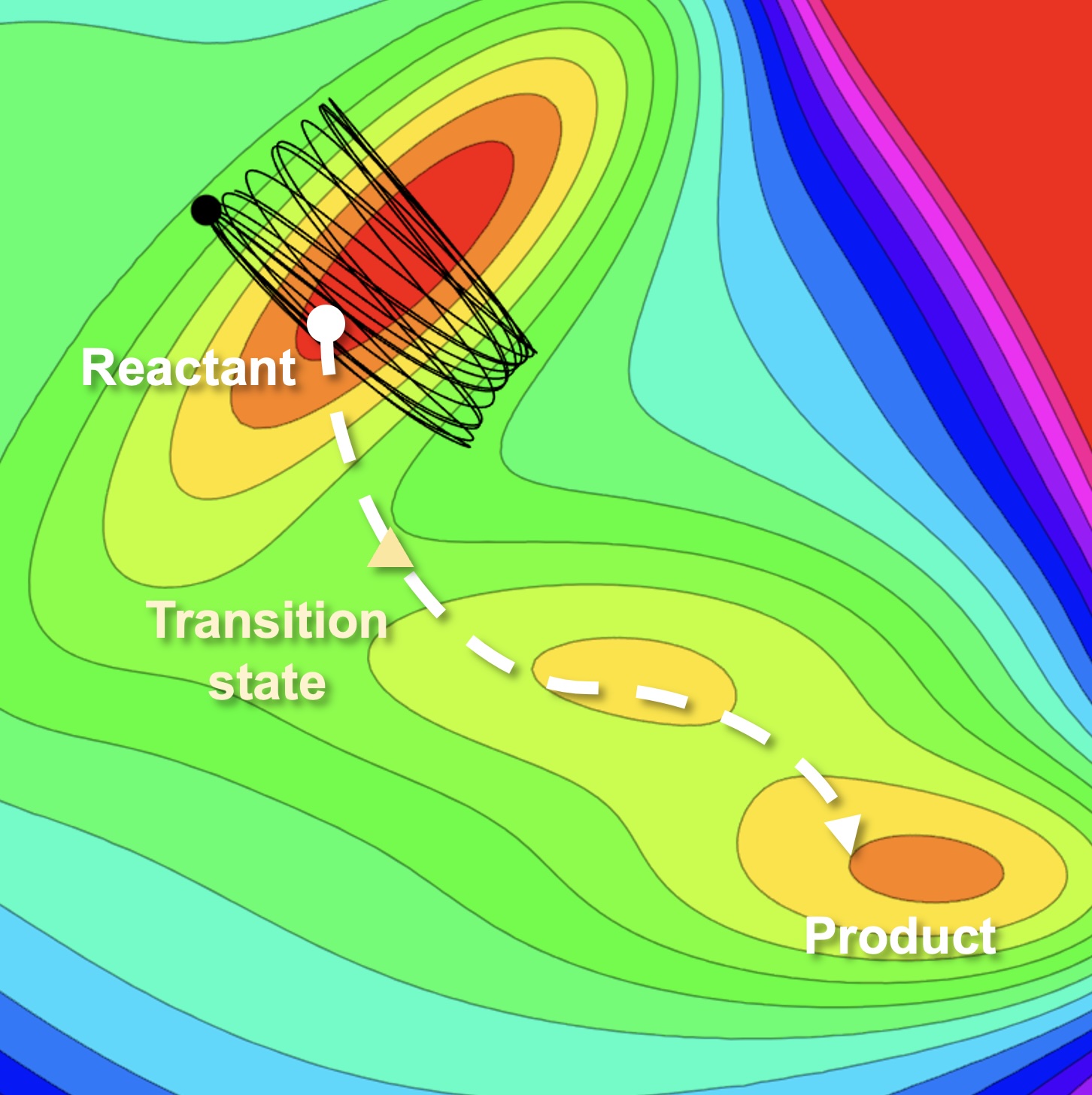}
\caption{Trajectories on a representative potential energy surface. The contour plot represents the energy landscape, with the color gradient indicating various energy levels. Trajectories are usually confined to regions near the minima, reflecting the system's preference for low-energy states close to or at equilibrium.}
\label{fig:mpb}
\end{figure}
Addressing the need for clarity, Christensen et al. revisited the potential energies and forces of the MD17 dataset, recalculating them using the PBE density functional with the def2SVP basis set and enhanced grid precision \cite{christensen2020role}. This effort led to the creation of the rMD17 dataset, which has since been widely adopted in NFF studies \cite{batzner20223,batatia2022mace}. Nonetheless, it is crucial to note the limitations of the PBE functional and def2SVP basis set for simulating accurate chemical reactions. While these computational tools can produce a continuous PES that varies with nuclear configuration, their ability to yield accurate results for chemically complex reactions — especially those involving bond breaking and formation — is often questioned. Despite these concerns, the MD17 and its refined counterpart, rMD17, are still considered to be well-behaved datasets for benchmarking purposes within certain constraints.

\begin{figure}[ht]
\centering
\includegraphics[width=\linewidth]{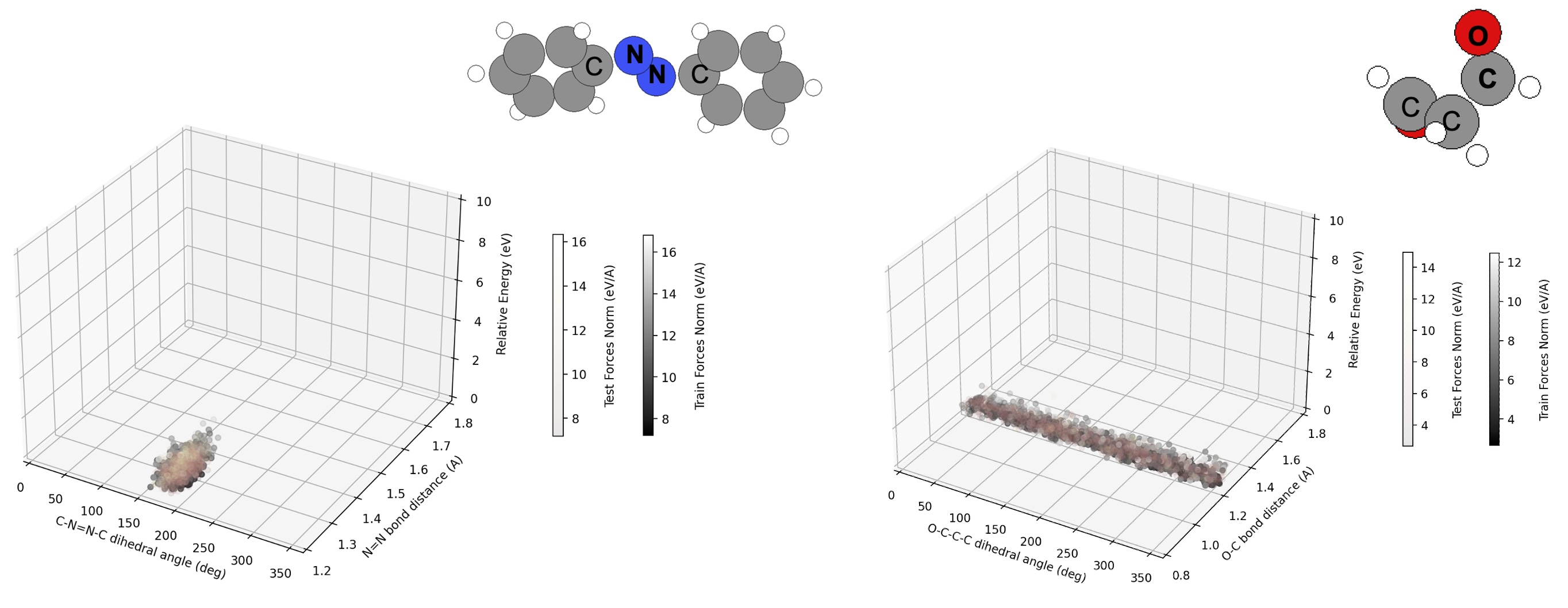}
\caption{Illustration of training and testing sets using the reference split indices for azobenzene and malonaldehyde datasets in rMD17. The X-axis depicts dihedral angles (marked by 'C', 'N', and 'O'), the Y-axis denotes bond distances (highlighted by bold letters), and the Z-axis shows relative energy. Training and testing samples are differentiated by color, correlating to force norms. Note that training samples overlap with testing ones.}
\label{fig:md17_split}
\end{figure}

Adiabatic molecular dynamics datasets generated at low energy range are inherently limited in their sampling diversity and may not benefit fully from techniques such as uniform sampling and cross-validation. This is particularly true for adiabatic AIMD simulations, where initial low-energy conditions substantially constrain the nuclear configuration space. This limitation results in trajectories that predominantly occupy the reactant zone of the PES, as depicted in Figure \ref{fig:mpb}.

To evaluate the breadth of configurations in the MD17 and rMD17 datasets, we conducted an analysis focused on internal coordinate distributions for azobenzene (C-N=N-C dihedral angle and the N=N bond length) and malonaldehyde (C-C-C=O dihedral angle and the C=O bond length). These distributions, along with the corresponding relative electronic potential energies and force norms, are illustrated in Figure \ref{fig:md17_split}. The visual representation confirms that the internal coordinates distribution is notably narrow. Consequently, we observe a significant overlap between the training and testing samples within these datasets. Such overlap raises concerns about potential data leakage, which could inadvertently lead to overly optimistic results in benchmarking studies, as discussed in the literature \cite{batatia2022design,batatia2022mace,batzner20223,schutt2021equivariant,liu2021spherical}. The findings underscore the need for datasets that encompass a more diverse and extensive sampling of the PES to ensure robust and reliable benchmarks for NFF models.

\subsection*{Dataset Requirement}

In classical MD and adiabatic AIMD simulations, chemical reactions are characterized by the system's transition across different minima on the PES. These transitions correspond to changes in electronic potential energy as the system moves through various nuclear configurations. Systems naturally tend to follow the path of least resistance, referred to as the reaction pathway. To develop accurate NFFs, two fundamental elements are required: a comprehensive quantum chemical dataset that captures the full range of molecular transformations from various regions, and an advanced machine learning model with the capacity to interpolate and extrapolate across the PES. The figure below illustrates typical trajectories on a PES. It's evident that trajectories tend to be localized around the ground state minima.

% \begin{figure}[ht]
% \centering
% \includegraphics[width=0.3\linewidth]{figs/mpb_2.jpg}
% \caption{Trajectories on a representative potential energy surface. The contour plot represents the energy landscape, with the color gradient indicating various energy levels. Trajectories are usually confined to regions near the minima, reflecting the system's preference for low-energy states close to or at equilibrium.}
% \label{fig:mpb}
% \end{figure}

In contrast, datasets derived from nonadiabatic dynamics simulations are particularly valuable as they provide a more diverse array of nuclear configurations, going beyond the limitations of low energy adiabatic AIMD. These enriched datasets allow for the exploration of PES regions that are critical for understanding complex chemical processes, which are often not adequately represented in low energy adiabatic simulations.

\subsection*{Summary}
In summary, the xxMD dataset developed in current work includes four molecular systems: azobenzene, malonaldehyde, stilbene, and dithiophene, with crucial geometries along their reaction pathways illustrated in Figure \ref{fig:xxmd_1}. Notably, azobenzene and malonaldehyde are also part of the MD17 and rMD17 datasets, allowing for direct comparison.

The geometries are sampled from nonadiabatic dynamics. The potential energies and gradients, i.e. forces, for the first three singlet electronic states at the state-averaged complete active state self-consistent field (SA-CASSCF) level of theory\cite{CASSCF} are included in xxMD-CASSCF dataset. In addition, spin-polarized KS-DFT with M06 functional\cite{zhao2008m06} calculations are performed on the same geometries as in xxMD-CASSCF dataset, the resulting ground singlet electronic state potential energies and gradients are included in the xxMD-DFT dataset. Therefore, the xxMD datasets developed in current work involve a multi-state dataset - xxMD-CASSCF dataset, and a single-state dataset - xxMD-DFT dataset. 

\section*{Method}

For our xxMD dataset, we employ the trajectory surface hopping (TSH) semiclassical nonadiabatic dynamics algorithm \cite{Tully1971,Truhlar1983,barbattiTSH} with SA-CASSCF electronic theory.\cite{CASSCF} The SA-CASSCF is a multireference electronic structure theory that provides qualitatively correct description of strong correlation - which are critical for deformed geometries and conical intersections, while the linear response time dependent Kohn-Sham density function approximations failed qualitatively.\cite{doubleexcitationinCI,DFTDA} We ensured that only energy-conserving trajectories were sampled. The size of the data samples is detailed in Table \ref{tab:merged} in supplementary material.

Nevertheless, to ensure compatibility with prevalent datasets like MD17, we also computed single-point spin-polarized KS-DFT (or unrestricted KS-DFT) values. These calculations employ the M06\cite{zhao2008m06} exchange-correlation functional — a notably superior meta-GGA functional relative to PBE. This dual approach culminates in two datasets: xxMD-CASSCF and xxMD-DFT. The former captures potential energies and forces across the first three electronic states for azobenzene, dithiophene, malonaldehyde, and stilbene. The latter provides recomputed ground-state energy and force values, anchored on the same trajectories. All computational details are described in supplementary information section G Computational details. Notice that SA-CASSCF PESs can be more complicated than DFT surfaces due to more complicated electronic structure algorithm from SA-CASSCF, i.e. choice of active space. Both xxMD datasets are structured via a temporal split method, partitioning training and testing data based on trajectory timesteps. We want to emphasize that xxMD datasets do not involve nonadiabatic coupling vectors (NACs) for two reasons: first, the advances in the field of nonadiabatic dynamics have enabled NAC-free nonadiabatic dynamics simulations, for example, curvature-driven dynamics.\cite{kappa1,kappa2,kappa3,kappa4,kappa5} Second, the purpose of the current work is to provide a database which includes a wide nuclear configuration space for which the energies and gradients of multiple electronic states are available. Therefore, the machine learning force field models can be tested against each surfaces. We note that an appropriate fit of a coupled PESs with multiple electronic states for a single system requires diabatic representation, which is beyond the discussion of the current work. \cite{shu2020diabatization,shu2021permutationally,diabaticreview} 

We evaluated six message-passing NFF models on the xxMD datasets: SchNet\cite{schutt2017schnet}, DimeNet++ (DPP)\cite{gasteiger2020fast}, SphereNet (SPN)\cite{liu2021spherical}, NequIP\cite{batzner20223}, Allegro\cite{musaelian2023learning}, and MACE\cite{batatia2022mace}. Each model was mostly used with its default parameters, and in line with convention, we trained the NFFs emphasizing more on force losses. While hyperparameter optimization could potentially improve performance (See Supplementary Information for an example), it remains outside the scope of this study. Therefore, the presented results might not showcase the absolute best performance for each model. Given our observations, we encourage researchers aiming to apply NFFs in practical scenarios to conduct rigorous re-benchmarks tailored to their specific chemical systems and objectives. 

Temporal splitting was chosen over random splitting to partition the xxMD datasets. This method involves dividing time-series data based on timesteps, reserving a specific range for testing and applying a 50:25:25 split for training, validation, and testing sets. Such a split allows for a rigorous assessment of a model’s ability to predict unexplored areas of the PES. This is highlighted in Figure \ref{fig:msd}, where deviations in trajectories over time emphasize the datasets' capability to challenge and evaluate the extrapolative power of NFFs. However, it is possible to use random splitting on xxMD datasets considering the wide coverage of conformation space. 

\section*{Data Records}

The xxMD-CASSCF and xxMD-DFT datasets have been made publicly available on GitHub at the following URL: \url{https://github.com/zpengmei/xxMD}; and on Zenodo at the following URL: \url{https://doi.org/10.5281/zenodo.10393859}.\cite{xxMDdataset} These datasets are stored in compressed archives, each containing pre-split extended XYZ format files based on temporal information. The files have been processed using the Atomic Simulation Environment (ASE) software package, as documented in the reference \cite{ase-paper}. The GitHub repository is structured into two main directories, each corresponding to one of the datasets: xxMD-CASSCF and xxMD-DFT.

Within each directory, data is further organized into subdirectories named after the four molecules studied: malonaldehyde, azobenzene, stilbene, and dithiophene. Each molecule's subdirectory contains the associated dataset files. Notably, the xxMD-CASSCF dataset includes an additional subdirectory structure that segregates the state-specific data for the first three electronic states.

\section*{Technical Validation}

\subsection*{Dynamic properties}

\begin{figure}
\centering
\includegraphics[width=0.8\linewidth]{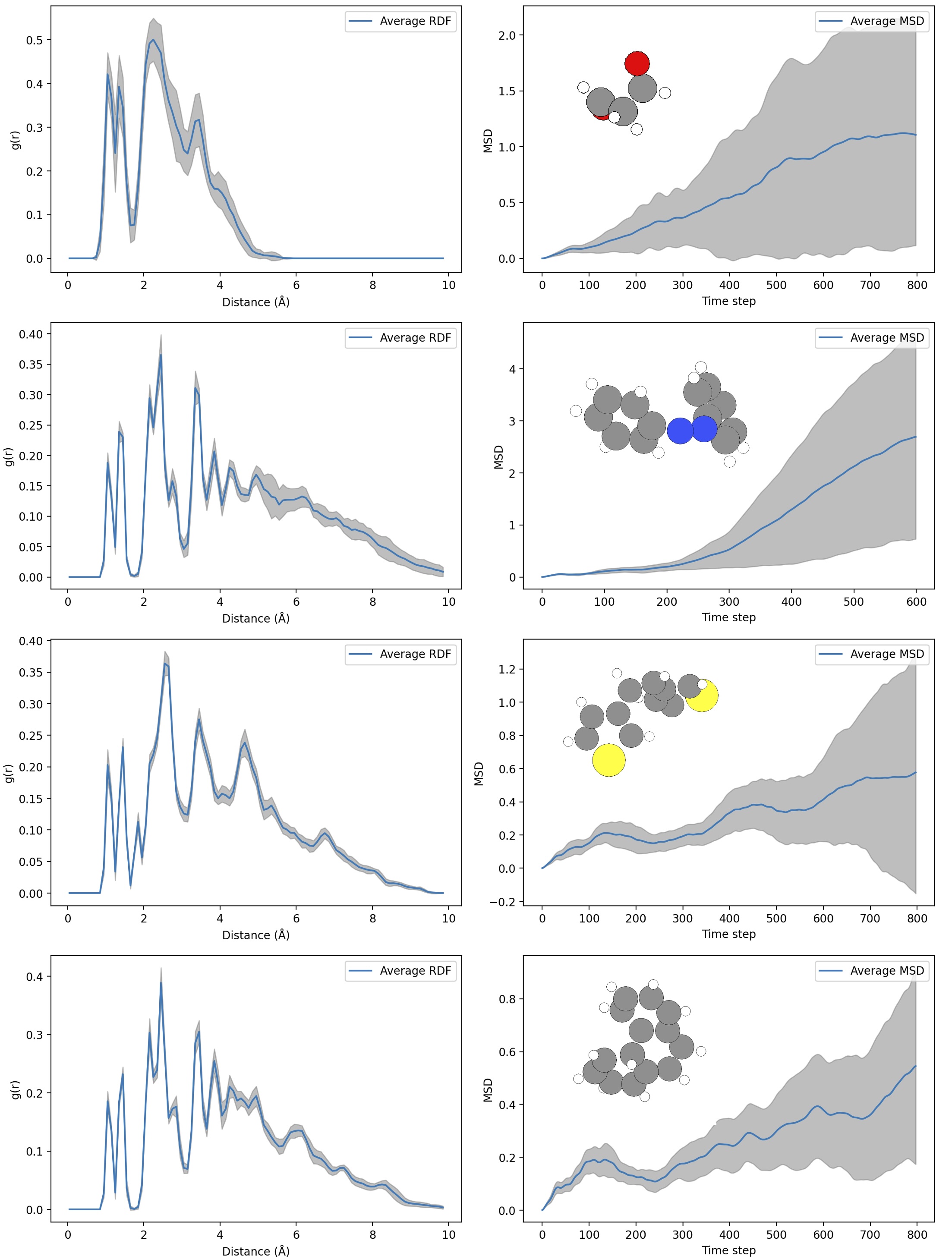}
\caption{Comparison of Average RDFs and MSDs Across Multiple Trajectories. Each row corresponds to a group of trajectories, with RDF on the left (indicating particle density as a function of distance) and MSD on the right (showing particle displacement over time). Shaded regions represent standard deviations.}
\label{fig:msd}
\end{figure}

Through the ensemble-averaged radial distribution function (RDF) and mean square displacement (MSD), the xxMD datasets exhibit a comprehensive sampling of the nuclear configuration space, surpassing that observed in MD17. Illustrated in Figure \ref{fig:msd}, the RDF and MSD track nuclear configurations over time, offering insights into the spatial distribution and mobility of particles, respectively. The RDF measures the likelihood of particle presence at varying radial distances from a reference point, whereas the MSD quantifies the average squared distance that molecules travel over a time interval.

The pronounced shifts in nuclear configurations captured by nonadiabatic dynamics in the xxMD datasets, as reflected in the dynamic breadth of the RDF and MSD, underline the enhanced diversity of PES regions sampled. Consequently, the complexity of mastering the PESs for molecules in the xxMD dataset is expected to be significantly elevated, presenting a robust challenge for the accuracy of NFFs.

\subsection*{Benchmarks on xxMD-CASSCF and xxMD-DFT datasets}
We picked six representative equivariant NFFs to benchmark. The hyperparameters and training details of models are described in the supplementary information. We used a weighted loss of 1:1000 on energy and forces. We stress that our purpose is not to perform an extensive comparison of models over multiple choices of hyperparameters. Rather, we limit ourselves to showing the performance of the models in the default configurations.

We first evaluate the regression precision of all models on the first three electronic states, which are labeled as S$_0$, S$_1$, and S$_2$ respectively (Label S denotes the singlet spin state which is a widely used notation in photochemistry) by using the temporal splitting approach for data in xxMD-CASSCF dataset. The MAE of the predictive energies and forces for test sets are shown in Table \ref{tab:results_time_cas}. Similarly, we present such results of using xxMD-DFT datasets in Table \ref{tab:results_time_dft}. Additional results on the validation sets are available in the supporting information. Note that validation sets depict the nuclear configurations that are closer to the training sets due to the temporal splitting. Therefore, the MAE shown in validation sets are in general lower than that for test sets. 

\begin{table}[ht]
\centering
\caption{Comparison of predictive MAE of energy(E, meV) and forces(F, meV/A) on hold-out testing set for different models on temporally split xxMD-CASSCF datasets and tasks. }
\begin{tabular}{lcccccccc}
\toprule
\textbf{Dataset} & \textbf{State} & \textbf{Task} & \textbf{MACE} & \textbf{Allegro} & \textbf{NequIP} & \textbf{SchNet} & \textbf{DPP} & \textbf{SPN} \\
\midrule
Azobenzene & S$_0$ & E & 527 & \textbf{437} & 870 & 648 & 528 & 493 \\
           &    & F & \textbf{63}  & 82  & 76  & 156 & 102  & 96 \\
           & S$_1$ & E & 599 & 524 & 1160 & 619 & 497  & \textbf{494} \\
           &    & F & \textbf{78}  & 98 & 85   & 157  & 91  & 88 \\
           & S$_2$ & E & 881 & 783 & 1957 & 894 & 837  & \textbf{831} \\
           &    & F & \textbf{191} & 216 & 215  & 284 & 224  & 231 \\
\midrule
Dithiophene & S$_0$ & E & 293 & 296 & 295 & 306  & 295 & \textbf{290} \\
            &    & F & \textbf{14}  & 31  & 21  & 94  & 30 & 31 \\
            & S$_1$ & E & 205 & 211 & 224 & 217  & \textbf{204} & 205 \\
            &    & F & \textbf{37}  & 81  & 49  & 103  & 41 & 44 \\
            & S$_2$ & E & 246 & 255 & 259 & 262  & \textbf{244} & 246 \\
            &    & F & 52  & 10 & 70  & 121  & \textbf{51} & 54 \\
\midrule
Malonaldehyde & S$_0$ & E & 530 & 443 & 770 & 515 & 452 & \textbf{442} \\
              &    & F & \textbf{105} & 142 & 166 & 220 & 138 & 137 \\
              & S$_1$ & E & 528 & \textbf{458} & 1227 & 482 & 482 & 462 \\
              &    & F & 164 & 189 & 189  & 260 & 165 & \textbf{161} \\
              & S$_2$ & E & 679 & \textbf{528} & 159 & 653 & 610 & 615 \\
              &    & F & 276 & 307 & 309  & 353 & 251 & \textbf{238} \\
\midrule
Stilbene & S$_0$ & E & 538 & 544 & 529 & 604 & \textbf{519} & 544 \\
         &    & F & \textbf{72}  & 87  & 112 & 191 & 91 & 114 \\
         & S$_1$ & E & 391 & 353 & 370 & 424 & \textbf{313} & 352 \\
         &    & F & \textbf{58}  & 66  & 85  & 142 & 88  & 93 \\
         & S$_2$ & E & 604 & 669 & 674 & 678 & 550  & \textbf{529} \\
         &    & F & \textbf{117} & 142 & 178 & 259 & 148  & 159 \\
\bottomrule
\end{tabular}

\label{tab:results_time_cas}
\end{table}

\begin{table}[ht]
\centering
\caption{Comparison of predictive MAE of energy(E, meV) and forces(F, meV/A) on hold-out testing set for different models xxMD-DFT datasets and tasks with temporal split.}
\begin{tabular}{lccccccc}
\toprule
\textbf{Dataset} & \textbf{Task} & \textbf{MACE} & \textbf{Allegro} & \textbf{NequIP} & \textbf{SchNet} & \textbf{DPP} & \textbf{SPN} \\
\midrule
Azobenzene & E & 292 & \textbf{174} & 1754 & 722 & 300 & 260 \\
           & F & \textbf{85}  & 110 & 129  & 283 & 173 & 168 \\
\midrule
Stilbene   & E & \textbf{315} & 332 & 647  & 397  & 439 & 477 \\
           & F & \textbf{149} & 189 & 156  & 291  & 162 & 168 \\
\midrule
Malonaldehyde & E & 190 & \textbf{151} & 244  & 360  & 179 & 185 \\
              & F & \textbf{166} & 210 & 227  & 394  & 257 & 255 \\
\midrule
Dithiophene & E & 100 & 103   & 243  & 323 & \textbf{61} & 76 \\
            & F & \textbf{51}  & 75   & 101  & 177 & 74  & 90 \\
\bottomrule
\end{tabular}
\label{tab:results_time_dft}
\end{table}

\subsection*{Comparison with existing datasets}

In this section, we analyze model behavior for two molecules, namely azobenzene and malonaldehyde. These two molecules are both available in xxMD and (r)MD17 datasets. Benchmarks for (r)MD17 reveal that the accuracy of MACE, NequIP, and SPN exceeds that of traditional electronic structure methods\cite{batatia2022mace,batzner20223,liu2021spherical,mardirossian2017thirty}. It's essential to note that typical errors for KS-DFT in predicting relative transition state energy can be several kcal/mol. For instance, the MAEs of HTBH38 (Hydrogen transfer barrier heights) and NHTBH38 (non-Hydrogen transfer barrier heights) databases are about 9.1 kcal/mol for PBE and 2.4 kcal/mol for M06. Thus, an NFF fitting error below 50 meV would surpass the accuracy of modern density functional calculations. However, such claims are pertinent mainly to ground state potential energies, given that excited state calculations are often less precise. Therefore, given the reported MAEs, these NFF models perform admirably on (r)MD17 datasets.

However, this conclusion might be deceiving. Previous discussions highlight the constrained nuclear configuration space in MD17 and rMD17. A comparative analysis of MAEs for the six NFF models on azobenzene and malonaldehyde from xxMD-DFT and (r)MD17 is presented in Table \ref{tab:model_sum}. Literature-derived MD17/rMD17 results indicate that all models used 1,000 training samples\cite{batatia2022mace,batzner20223,liu2021spherical}. Predictably, the predictive prowess of NFF models diminishes when applied to the xxMD dataset. 

The differences of MAEs for a same NFF model for rMD17 and xxMD come from two aspects, namely, the differences in dataset, and the differences in splitting method. The xxMD datasets contain much more complex nuclear configurations than (r)MD17. For the splitting method, one can have either random splitting or temporal splitting. For certain purposes, for example, if one uses the trajectory data to construct a global PES for the system, random splitting would be a good approach. For purpose of extended trajectory simulation with existing trajectory data, temporal splitting may be favored. Because the ultimate goal is to look for unknown chemical events that may not be observable from short trajectory simulations. In that spirit, we use temporal splitting in the current work. For the purpose of extended trajectory simulation, random splitting, which has been used to test against (r)MD17 dataset, means a severe leakage of future information. In practice, if we would like to model a chemical reaction, it would be impractical to manually sample every relevant region on the potential energy surfaces. Therefore, it is a desired property for an NFF model has the capability of physical extrapolation to some extent. Physical extrapolation is achieved in several models, for examples, reactive force field,\cite{reaxff} and parametrically managed activation function.\cite{pmddnn}

\begin{table}[h]
\centering
\caption{Comparison of predictive MAE on hold-out testing sets of NFF models on azobenzene and malonaldehyde in (r)MD17 and xxMD-DFT datasets. (r)MD17 benchmarks with 1,000 samples are taken from \cite{batatia2022mace,liu2021spherical,schutt2017schnet}.}
\begin{tabular}{llccccccc}
\toprule
\textbf{Molecule} & \textbf{Dataset} & \textbf{Task} & \textbf{MACE} & \textbf{Allegro} & \textbf{NequIP} & \textbf{SchNet} & \textbf{DPP} & \textbf{SPN} \\
\midrule
Azobenzene & rMD17 & E & 1.2 & 1.2 & 0.7 & N/A & N/A & N/A \\
           &       & F & 3.0 & 2.6 & 2.9 & N/A & N/A & N/A \\
           & xxMD  & E & 292 & 174 & 1754 & 722 & 300 & 260 \\
           &       & F & 85 & 110 & 129 & 283 & 173 & 168 \\
\midrule
Malonaldehyde & (r)MD17 & E & 0.8 & 0.6 & 0.8 & 5.6 & 4.5 & N/A \\
              &       & F & 4.1 & 3.6 & 5.1 & 28.6 & 16.6 & 7.5 \\
              & xxMD  & E & 190 & 151 & 244 & 360 & 179 & 185 \\
              &       & F & 166 & 210 & 227 & 394 & 257 & 255 \\
\bottomrule
\end{tabular}

\label{tab:model_sum}
\end{table}

The effectiveness of NFF models largely depends on the datasets they are benchmarked against. Historically, the (r)MD17 datasets have been the gold standard for this purpose. However, our study highlights the potential shortcomings of relying solely on (r)MD17 datasets. Given that they primarily capture a narrow nuclear configuration space from low energy ground state AIMDs, they fall short of encompassing the holistic nuclear configuration pertinent to chemical reactions. Training NFF models on such datasets can be somewhat trivial and could result in misleading conclusions about their true capabilities. For instances, computational chemists have a long history of using system specific force fields, which can be easily developed by computing a hessian at the ground state equilibrium geometry.\cite{JOYCE,quickFF}

To address this gap, we introduced the xxMD dataset, derived from nonadiabatic dynamics trajectories. The xxMD dataset offers a comprehensive representation of the nuclear configuration space, encapsulating the reactant, transition state, product, and conical intersection regions of PESs. Its inclusion of several low-lying excited state potential energy surfaces underscores its importance and the challenges it presents for NFF model development. Our benchmarks of prevailing NFF models on the xxMD dataset have revealed pronounced difficulties. Utilizing default hyperparameters, the chosen NFF models struggled to offer quantitatively or even qualitatively accurate force field models for specific systems. We anticipate that our findings will galvanize the community towards pioneering more advanced NFF models better equipped to study intricate chemical reactions.

\section*{Code availablity}
Nonadiabatic dynamics are performed with Surface Hopping with Arbitrart Coupling (SHARC) code, which is available at \url{https://github.com/sharc-md/sharc}. SchNet, DimeNet++ and SphereNet are available as implemented in the Dive Into Graphs package (\url{https://github.com/divelab/DIG.git}). NequIP package is available at \url{https://github.com/mir-group/nequip.git}. Allegro package is available at \url{https://github.com/mir-group/allegro}. MACE package is available at \url{https://github.com/ACEsuit/mace.git}. All packages are up-to-date at the data of the publication. All the trainings are done with single precision float format. SchNet, DPP and SPN models are initialized using the default hyperparameters shipped with the packages. Allegro hyperameters can be found at \url{https://github.com/mir-group/allegro/blob/main/configs/example.yaml}, NequIP hyperparameters are mainly \url{https://github.com/mir-group/nequip/blob/main/configs/example.yaml}, MACE hyperparameters are mainly \url{https://github.com/ACEsuit/mace}. Since Dive Into Graphs package doesn't implement the scale and shift of the energy, we manually rescaled the energy by substracting the energy of the configuration with the lowest potential energy. 

\section*{Acknowledgement}

Authors acknowledge helpful discussion with Prof. Erik H. Thiede. Early versions of the draft employs writing advice from OpenAI's ChatGPT. J. L. is supported in part by International Business Machines (IBM) Quantum through the Chicago Quantum Exchange, and the Pritzker School of Molecular Engineering at the University of Chicago through AFOSR MURI (FA9550-21-1-0209).

\section*{Contribution}

Z.P. conceived and conceptualized the idea and designed the experiment. Z.P. and Y.S. performed the experiments and analyzed the data. Y.S. and Z.P. provided the computational resources. Y.S. and J.L. participated discussion with Z.P. Z.P. and Y.S. drafted the manuscript and all authors reviewed and agreed with the manuscript.

\section*{Competing interests}
The authors declare no competing interests.

\bibliography{sample}
\end{document}

\appendix
\setcounter{page}{1}

\setcounter{figure}{0}    
\setcounter{table}{0}
\renewcommand{\thefigure}{S\arabic{figure}}
\renewcommand{\thetable}{S\arabic{table}}

% Start a new ToC for the SI
\startcontents[sections]
\section*{Supplementary Information Table of Contents}
\printcontents[sections]{l}{1}{\setcounter{tocdepth}{2}}

% Create the SI ToC
% \listofappendices

\section{Preliminaries of dynamics}

In the realm of quantum mechanics, the behavior of nuclei is ideally described by the time-dependent Schrödinger equation. Yet, practical computation limits restrict nuclear quantum dynamics simulations to small systems with just 5 or 6 atoms. Consequently, in many cases, the nuclei are treated as classical particles. This premise paves the way for classical Molecular Dynamics (MD) and adiabatic Ab Initio Molecular Dynamics (AIMD), wherein the dynamics are propagated based on a \textbf{single electronic state}.

At the heart of classical MD is the Newtonian equation of motion:
\begin{equation}
m_i \frac{d^2 \mathbf{r}_i}{dt^2} = \mathbf{F}_i
\end{equation}
where \( m_i \) denotes the mass of atom \( i \), \( \mathbf{r}_i \) its position, and \( \mathbf{F}_i \) the force exerted on it. This force can be described as the negative gradient of the potential energy \( V \) at the atom's location:
\begin{equation}
\mathbf{F}_i = -\nabla V(\mathbf{r}_i)
\end{equation}
The ground state electronic potential energy, \( V(\mathbf{r}_i) \), in the absence of an external field, forms the basis for the PES. Classical force fields offer an analytical approximation of this energy based on nuclear configuration:
\begin{equation}
V(\mathbf{r}) = V_{\text{bond}}(\mathbf{r}) + V_{\text{angle}}(\mathbf{r}) + V_{\text{dihedral}}(\mathbf{r}) + V_{\text{non-bonded}}(\mathbf{r})
\end{equation}
This classical approximation often falls short under quantum mechanical scenarios, particularly during bond breaks, necessitating improvements in force field formulations. Upon electronic excitation, as observed in solar cells or photochemical reactions, nuclei confront electronic potentials beyond the ground state. Herein, dynamics involving multiple electronic states emerge. Nonadiabatic dynamics, particularly pertinent when energy levels soar, may either adopt the trajectory surface hopping method or the semiclassical Ehrenfest dynamics, depending on the specific conditions.

\section{Brief introduction of chosen neural force fields}
In this study, we picked six representative neural network architectures for NFF applications, namely, SchNet\cite{schutt2017schnet}, DPP\cite{gasteiger2020fast}, SPN\cite{liu2021spherical}, NequIP\cite{batzner20223}, Allegro\cite{musaelian2023learning} and MACE\cite{batatia2022mace}. In general, those approaches can be devided into two catagories based on the representation of the feature space. SchNet, DPP and SPN are the so-called scaler-based NFFs, while NequIP, MACE and Allegro are vector-based NFFs, as we summarized in Table \ref{tab:model_features}.

The key concept in SchNet is the continuous-filter convolution, which involves two steps: interaction and update. In the interaction step, the model calculates pairwise interaction features between all atoms based on their distances, using a set of radial bessel basis. The update step then uses these interaction features to update the atom-centered descriptors. In DPP, a higher-order feature, bond angle has been introduced to enhance the expressiveness of the neural network. DPP uses a concept called spherical functions to account for the directionality of the interactions between atoms. The DPP architecture uses 'interaction blocks' to propagate information through the molecular graph. Each interaction block consists of a radial and a spherical part. The radial part captures the distance-based interactions, similar to SchNet. The spherical part captures the angular interactions among any three atoms in the molecule, which is unique to DPP. As a continuation of the DPP, SPN further introduces another higher-order feature called dihedral angles among any four atoms in the molecule. These improvements are chemically-intuitive since bond lengths, angles, and dihedral angles are very common descriptors in classical force fields \cite{molecularthermo}. 

\begin{table}[ht]
\centering
\caption{Summary of models, their features, and the corresponding years of introduction.}
\begin{tabular}{lll}
\toprule
\textbf{Model} & \textbf{Feature} & \textbf{Year} \\
\midrule
SchNet      & Bond length                      & 2017 \\
DPP   & Bond length, Bond angle          & 2020 \\
SPN   & Bond length, Bond angle, Dihedral angle   & 2021 \\
NequIP      & SO(3) vector                     & 2021 \\
Allegro     & SO(3) vector                     & 2022 \\
MACE        & SO(3) vector                     & 2022 \\
\bottomrule
\end{tabular}
\label{tab:model_features}
\end{table}

\begin{figure}[ht]
\centering
\includegraphics[width=1\textwidth]{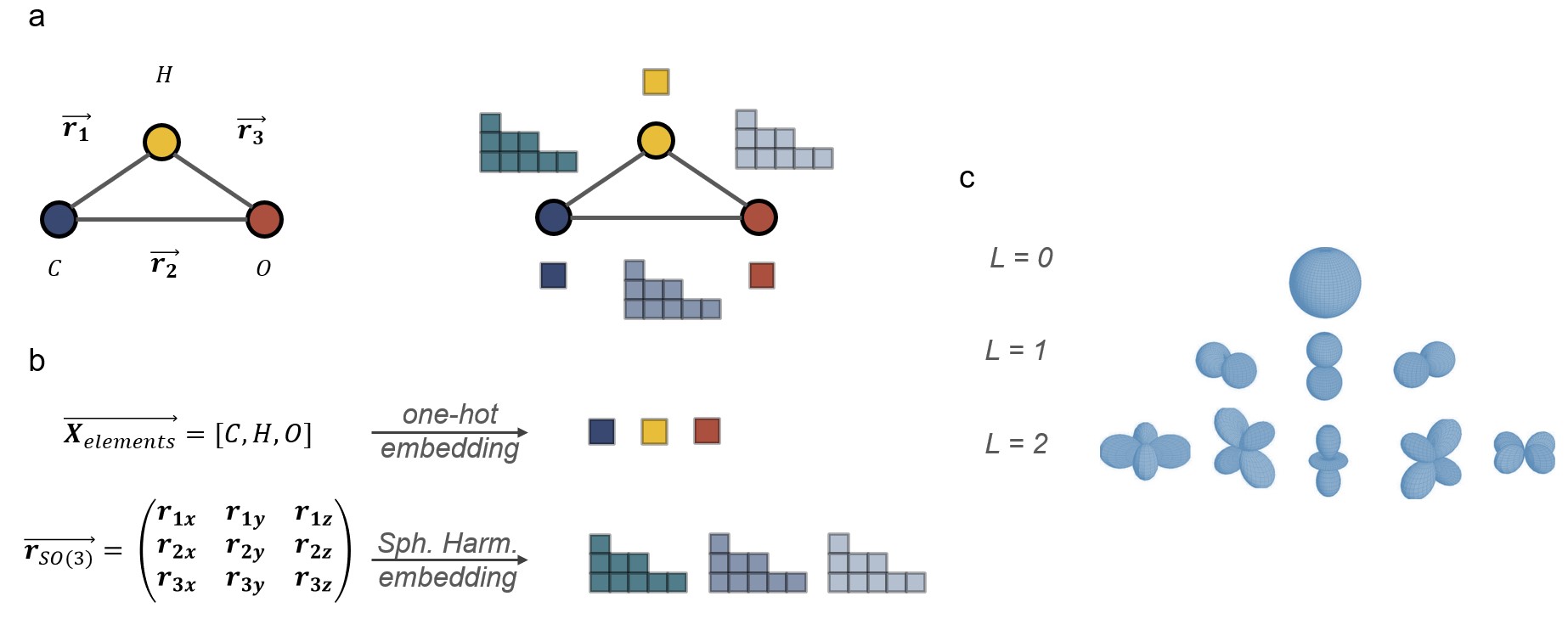}
\caption{(a) depicts three atoms with their relative displacement vectors. (b) illustrates the details of atomic embedding with E(3) equivariant activations based on spherical harmonics and one-hot encoding for chemical elements. (c) gives an illustration of spherical harmonics with quantum numbers $L=0, 1, 2$. The Clebsch-Gordan coefficients are used during the aggregation step to ensure rotational equivariance when combining activations with different irreps.}
\label{fig:eqnn}
\end{figure}

On the other hand, NequIP, Allegro, and MACE are examples of group equivariant NFFs that based on SO(3) relative displacement vectors between any two atoms in the molecule. These networks use the representation theory of the three-dimensional orthogonal group to construct neurons that obey equivariance with respect rotations and reflections of a  molecular system's pose. We visualize this concept in Figure \ref{fig:eqnn}. Atomic types are embedded as node features, and relative displacement vectors that contains the positional information are converted into activations that transform according to irreducible representations (irreps) of the orthogonal group. Nonlinearities for these activations are constructed using the tensor product, followed by applying the Clebsch-Gordon decomposition to convert the product back into irreducible components. 

\section{Timing}

\begin{figure}
    \centering
    \includegraphics[width=0.7\textwidth]{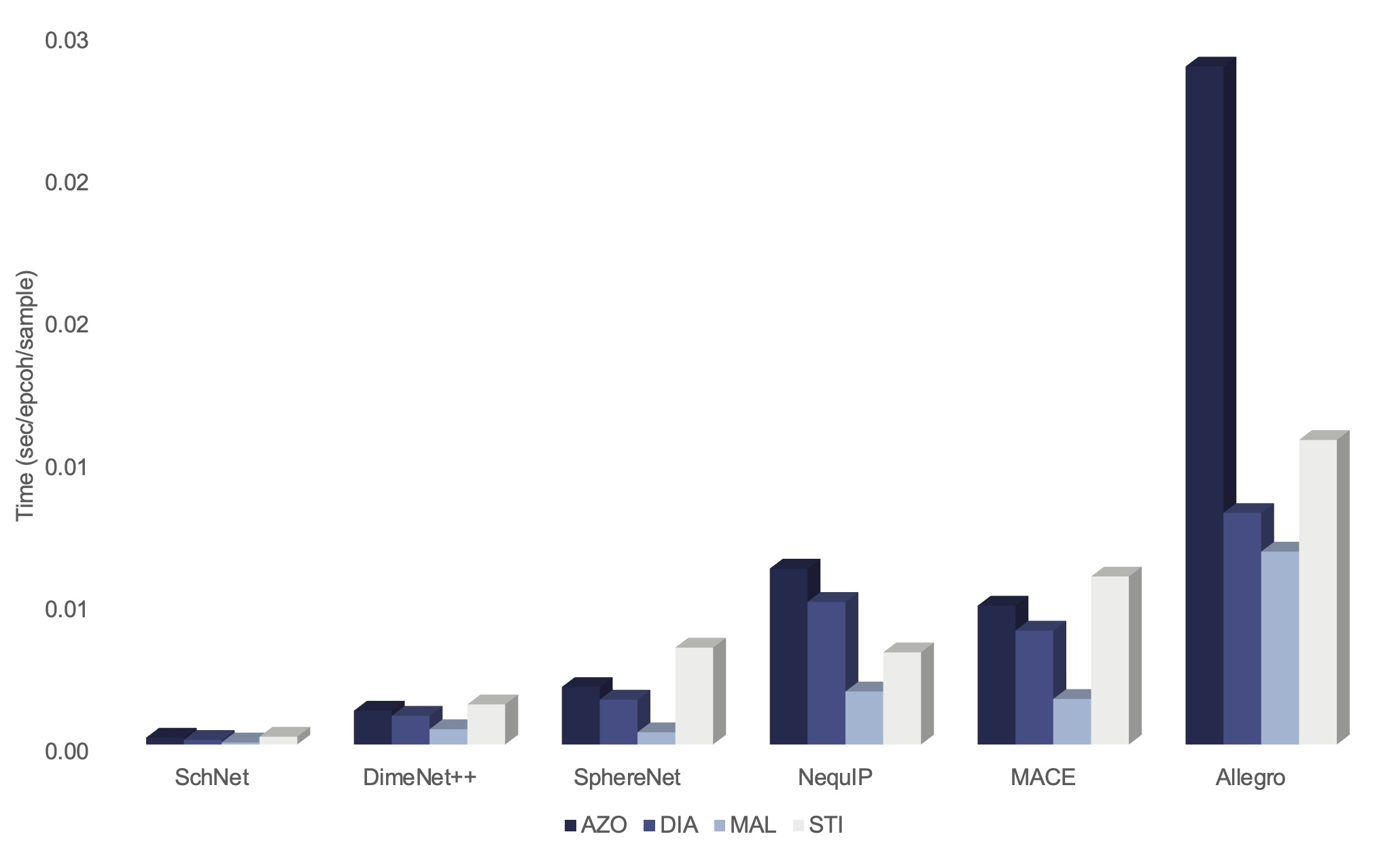}
    \caption{Comparison of average computational time for NFFs. The timing is specific to the chosen hyperparameters. All NFFs, except MACE, operate with single precision. Generally, group-equivariant NFFs are significantly more computationally expensive.}
    \label{fig:timing}
\end{figure}

In this practical view, we present a comprehensive analysis of the operational time of multiple NFFs examined in our study as illustrated in Figure \ref{fig:timing}. It is important to note that the specific runtime of each NFF model is contingent upon the chosen setup and hyperparameter selection. For example, the radius cutoff utilized for generating locally fully-connected graphs can yield varying numbers of edges and nodes in each mini-batch. Within our findings, we have diligently reported the time required for processing each sample in a mini-batch using the designated hyperparameters. Consequently, we emphasize that while we employed mostly default hyperparameters as a practical reference.

\section{Additional illustration of xxMD datasets}

\begin{figure}[ht]
\centering
\includegraphics[width=0.7\linewidth]{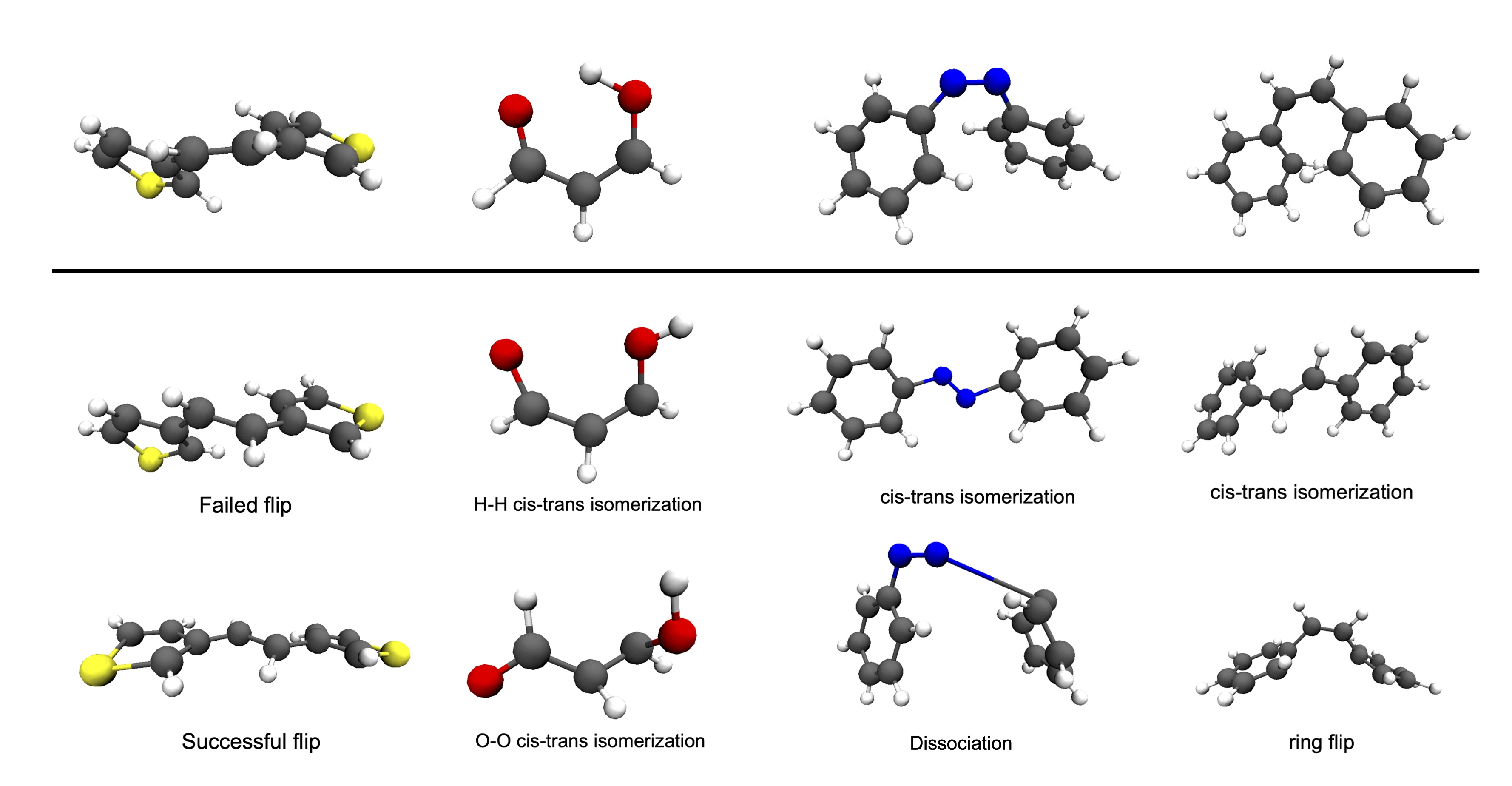}
\caption{Schematic representation of the photodynamic processes featured in the xxMD dataset.}
\label{fig:xxmd_1}
\end{figure}

\begin{figure}
\centering
\includegraphics[width=\textwidth]{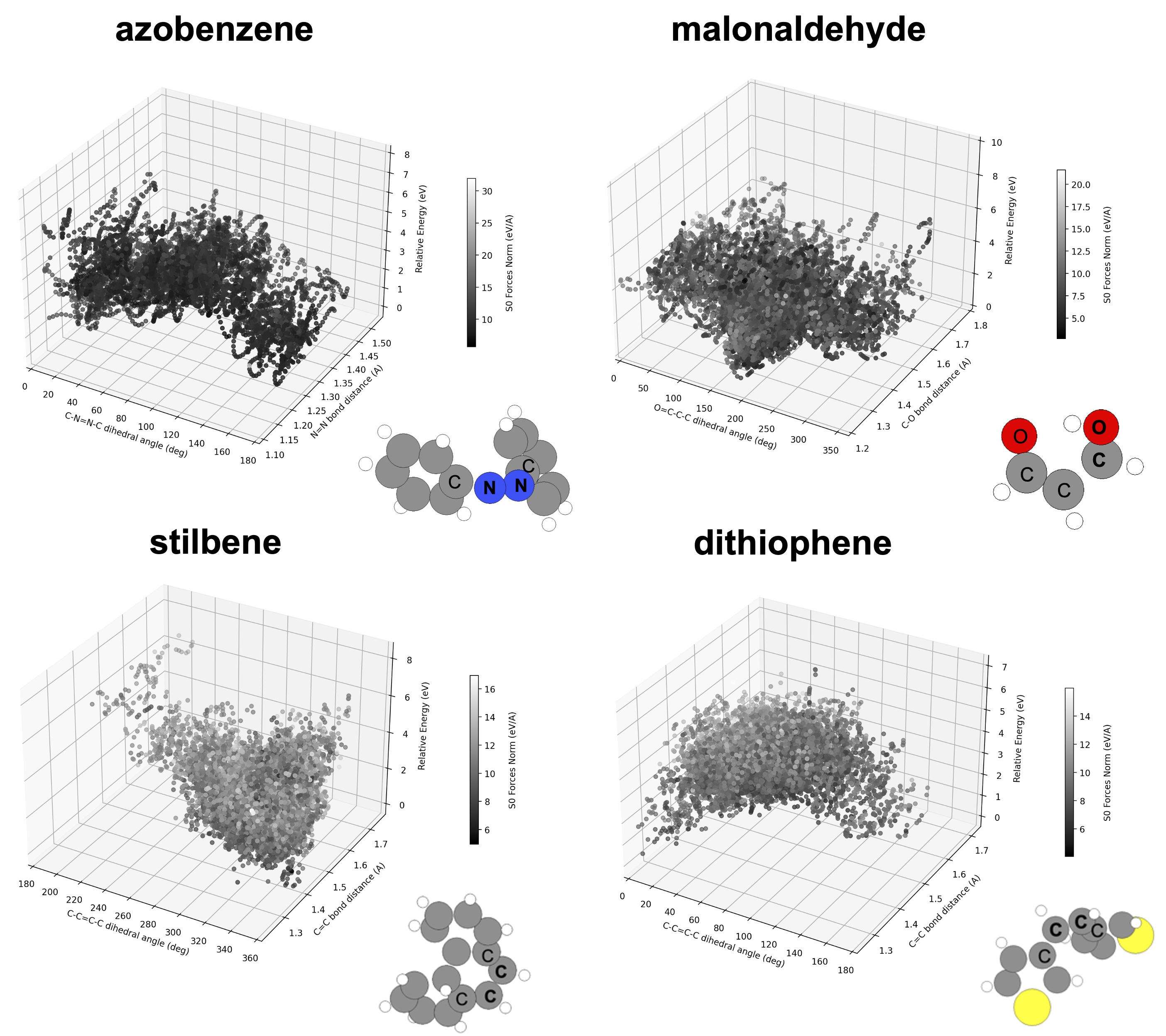}
\caption{Illustration of xxMD datasets using similar internal coordinates as MD17 analysis. Only the ground-state (in black/white color scheme) energies are visualized for clarity. This figure clearly indicates the breadth of the conformation space explored by using direct non-adiabatic dynamics as compared to MD simulation with room temperature.}
\label{fig:xxmd_1_3d}
\end{figure}

Here we provide addition illustration (Figure \ref{fig:xxmd_1_3d}) of the xxMD-CASSCF datasets with the ground-state energy and forces as the internal coordinate analysis of MD17. For azobenzene, the primary reaction path involves the cis-trans isomerization of the two phenyl groups along the N=N bond. For malonaldehyde, the reaction path involves either a H-H cis-trans isomerization occurs along the O=C bond or a O-O cis-trans isomerization occurs along the carbon skeleton. The reaction path of stilbene involves the cis-trans isomerization of the two phenyl rings along the C=C double bond and the flip of the phenyl rings to opposite directions. The reaction path of dithiophene is also the cis-trans isomerization of two five-member rings along the C=C double bond.
\newpage

\section{Benchmark results on the validation sets of xxMD-CASSCF and xxMD-DFT}

\begin{table}[ht]
\centering
\caption{Comparison of predictive MAE on validation set for different models on temporally split xxMD-CASSCF datasets and tasks. Energy(E) has the unit of meV, while forces(F) have the unit of meV/A.}
\begin{tabular}{lcccccccc}
\toprule
\textbf{Dataset} & \textbf{State} & \textbf{Task} & \textbf{MACE} & \textbf{Allegro} & \textbf{NequIP} & \textbf{SchNet} & \textbf{DPP} & \textbf{SPN} \\
\midrule
Azobenzene & $S_0$ & E & 527 & \textbf{367} & 530 & 682 & 526 & 494 \\
           &    & F & \textbf{50}  & 70  & 69  & 141 & 83  & 79 \\
           & $S_1$ & E & 474 & \textbf{308} & 869 & 483 & 478  & 428 \\
           &    & F & \textbf{67}  & 83  & 74   & 134  & 79  & 74 \\
           & $S_2$ & E & 864 & \textbf{742} & 1590 & 897 & 804  & 801 \\
           &    & F & \textbf{163} & 185 & 180  & 257 & 191  & 185 \\
\midrule
Dithiophene & $S_0$ & E & 300 & 295 & 304 & 302  & \textbf{286} & 287 \\
            &    & F & \textbf{10}  & 21  & 17  & 76  & 22 & 24 \\
            & $S_1$ & E & 259 & 208 & 226 & 219  & \textbf{206} & 208 \\
            &    & F & 65  & 78  & 46  & 101  & \textbf{33} & 36 \\
            & $S_2$ & E & 246 & 258 & 256 & 259  & \textbf{244} & 249 \\
            &    & F & 50  & 104 & 69  & 119  & \textbf{49} & 51 \\
\midrule
Malonaldehyde & $S_0$ & E & 488 & \textbf{386} & 583 &470 & 419 & 415 \\
              &    & F & \textbf{84} & 147 & 109 & 179 & 108 & 109 \\
              & $S_1$ & E & 507 & \textbf{406} & 828 & 469 & 446 & 451 \\
              &    & F & \textbf{144} & 184 & 168  & 233 & 147 & 145 \\
              & $S_2$ & E & 556 & \textbf{457} & 858 & 526 & 512 & 512 \\
              &    & F & 221 & 255 & 281  & 301 & 197 & \textbf{188} \\
\midrule
Stilbene & $S_0$ & E & 517 & 514 & \textbf{359} & 505 & 467 & 461 \\
         &    & F & 54  & 71  & \textbf{12} & 145 & 71 & 75 \\
         & $S_1$ & E & 322 & 293 & \textbf{262} & 351 & 294 & 316 \\
         &    & F & 38  & 45  & \textbf{20}  & 97 & 62  & 61 \\
         & $S_2$ & E & 494 & 505 & \textbf{377} & 596 & 486  & 473 \\
         &    & F & 80 & 98 & \textbf{31} & 176 & 104  & 106 \\
\bottomrule
\end{tabular}

\label{tab:results_time_cas_val}
\end{table}

\begin{table}[ht]
\centering
\caption{Comparison of predictive MAE on validation set for different models xxMD-DFT datasets and tasks with temporal split. Energy(E) has the unit of meV, while forces(F) have the unit of meV/A.}
\begin{tabular}{lccccccc}
\toprule
\textbf{Dataset} & \textbf{Task} & \textbf{MACE} & \textbf{Allegro} & \textbf{NequIP} & \textbf{SchNet} & \textbf{DPP} & \textbf{SPN} \\
\midrule
Azobenzene & E & 257 & 106 & 393 & 539 & 184 & \textbf{168} \\
           & F & \textbf{71}  & 98 & 119  & 248 & 150 & 140 \\
\midrule
Stilbene   & E & 190 & 200 & 161  & \textbf{156}  & 224 & 248 \\
           & F & \textbf{104} & 116 & 117  & 196  & 114 & 125 \\
\midrule
Malonaldehyde & E & 156 & \textbf{91} & 134  & 257  & 116 & 127 \\
              & F & \textbf{135} & 162 & 173  & 326  & 208 & 204 \\
\midrule
Dithiophene & E & 89 & 54   & 86  & 198 & \textbf{49} & 69 \\
            & F & \textbf{47}  & 59   & 81  & 158 & 61  & 78 \\
\bottomrule
\end{tabular}

\label{tab:results_time_dft_val}
\end{table}

\section{Addition experiment of hyperparameter tuning}

We would like to stress again, our purpose is to give a initial view of the datasets using common hyperparameters without tuning, and we don't aim to strictly test models listed. We left most hyperparameters unchanged as default, and uses a loss weight heavily focused on the forces following the literatures \cite{stocker2022robust,batatia2022mace,batzner20223}. However, users should carefully use the hyperparameters before applying to specific chemical problems. 

\begin{table}[ht]
\centering
\caption{Predictive MAE of energy (meV) and forces (meV/A) on the ground-state azobenzene in xxMD-CASSCF dataset using various loss weights and default MACE model. }
\label{tab:mace_loss_weights}
\begin{tabular}{@{}ccccc@{}}
\toprule
\textbf{}               & \multicolumn{2}{c}{\textbf{Testing}} & \multicolumn{2}{c}{\textbf{Validation}} \\ \midrule
\textbf{Loss E:F ratio} & \textbf{E}        & \textbf{F}       & \textbf{E}         & \textbf{F}         \\
\midrule
1000:1                  & 325               & 210              & 291                & 186                \\
100:1                   & 311               & 104              & 266                & 87                 \\
10:1                    & 338               & 72               & 327                & 58                 \\
1:1                     & 446               & 66               & 458                & 53                 \\
1:10                    & 516               & 64               & 524                & 50                 \\
1:100                   & 541               & 65               & 544                & 50                 \\
1:1000                  & 527               & 63               & 527                & 50                 \\ \bottomrule
\end{tabular}
\end{table}

We used a default MACE model and one subset of xxMD-CASSCF dataset and varied the weights on the energy and forces, and we found that by simply tuning this hyperparameter, MACE would perform noticeably differently. For instance, the regression accuracy on force is not improved and accuracy on energy deteriorates quickly when the weight on the force gradually increase from 1 to 1000. On the contrary, putting slightly more weights on the energy greatly improve the overall performance, as we laid out in Table \ref{tab:mace_loss_weights}. Thus, we would like to leave a note to future users that exploring the hyperparameter spaces is important. 
\section{Computational details}
\begin{figure}[ht]
\centering
\includegraphics[width=0.75\linewidth]{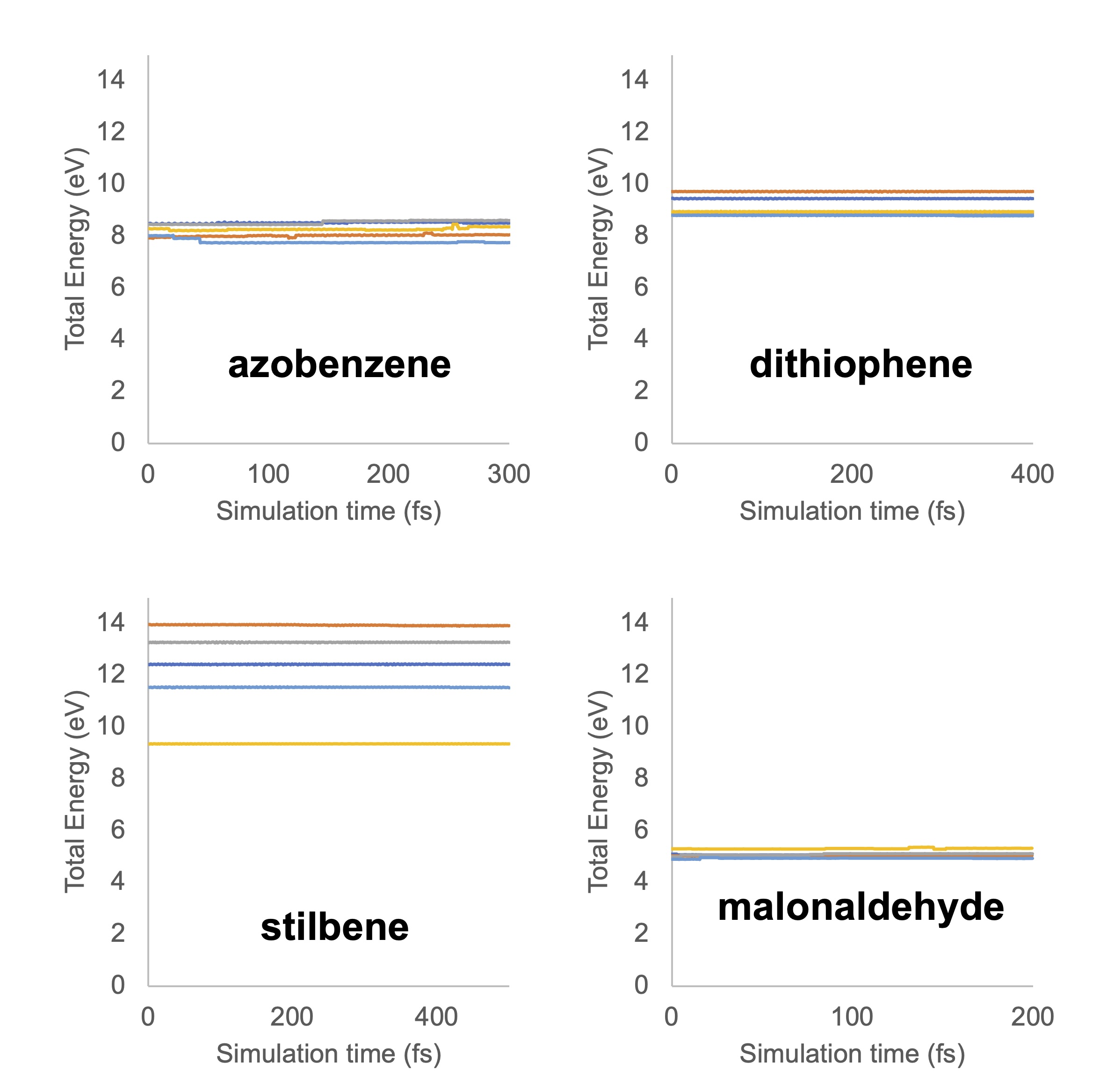}
\caption{Illustration of total energy conservation over the simulation time of trajectories in xxMD-CASSCF datasets. All trajectories follow the total energy conservation threshold.}
\label{fig:ec}
\end{figure}

The active space and basis set used for SA-CASSCF for all four molecules are shwon in Table \ref{tab:activespace}. The total number of trajectories simulated are vary, but finally selected number of points for each molecule in xxMD dataset is summarized in Table \ref{tab:merged} as well. These points are selected from energy conserving trajectories only, and we used the criteria for the total energy conservation as listed in Table \ref{tab:merged}. Therefore, all trajectories fail to conserve the total energy below the threshold are discarded. We show the total energy conservation in Figure \ref{fig:ec}
\begin{table}[ht]
\centering
\caption{Summary of the computational methods, number of samples used in direct non-adiabatic dynamics sampling for four molecules, and number of data points for all studied molecules. The number in the parenthesis indicates the number of active electrons and orbitals. The total energy conservation (Total E. Con.) criteria has a unit of eV.}
\begin{tabular}{l|lcc}
\toprule
\textbf{Molecule} & \textbf{Method} &\textbf{Total E. Con.}& \textbf{Num. of Samples}  \\
\midrule
Dithiophene & SA-CASSCF(10e,10o)/6-31g & 0.2 & 24769  \\
Azobenzene & SA-CASSCF(6e,6o)/6-31g & 0.6 & 8414  \\
Malonaldehyde & SA-CASSCF(8e,6o)/6-31g & 0.3 & 25568  \\
Stilbene & SA-CASSCF(2e,2o)/6-31g* & 0.2 & 27965  \\
\bottomrule
\end{tabular}
\label{tab:activespace}
\end{table}

\begin{table}[ht]
\centering
\caption{Summary of the number of samples used in direct non-adiabatic dynamics sampling for four molecules, and number of data points for all studied molecules.}
\begin{tabular}{l|cccc}
\toprule
\textbf{Molecule} & \textbf{Num. of Samples} & \textbf{Train} & \textbf{Valid} & \textbf{Test} \\
\midrule
Dithiophene & 24769 & 12400 & 6169 & 6200 \\
Azobenzene & 8414 & 4200 & 2114 & 2100 \\
Malonaldehyde & 25568 & 14000 & 6965 & 7000 \\
Stilbene & 27965 & 12800 & 6368 & 6400 \\
\bottomrule
\end{tabular}
\label{tab:merged}
\end{table}

\subsection{Dynamics}

 Initial conformations are generated by Wigner-Sampling of the optimized ground-state structure with the same level of electronic structure method. For each conformation, a single-point calculation is performed to acquire the energy of states without spin-orbit calculations. To select initial excited-states, the MCH representation of the Hamiltonian is used to simulate delta-pulse excitation based on excitation energies and oscillators strengths with an excitation window of 0.0 to 10.0 eV.

For azobenzene, we conducted 300 fs SHARC dynamics simulations with a time step of 0.5 fs. For dithiophene, we conducted 500 fs SHARC dynamics simulations with a time step of 0.5 fs. For malonaldehyde, we conducted 300 fs SHARC dynamics with a timestep of 0.25 fs. For stilbene, we performed 500 fs SHARC dynamics with a time step of 0.5 fs. Local diabatizatrion scheme was used to calculate the non-adiabatic coupling vectors by calculating the overlap matrix of wavefunctions between steps. Non-adiabatic coupling vectors are included in the gradient transformation. kinetic energy are adjusted by rescaling the velocity vectors during a surface hop. When the surface hop is refused due to insufficient energy, the velocity doesn't reflect at a frustrated hop. Default energy-based decoherence scheme was used for decoherence correction. The standard SHARC surface hopping probabilities was used as the surface hopping scheme. All gradients and non-adiabatic couplings of active states were calculated at each time step. For azobenzene, dithiophene, malonaldehyde, and stilbene the threshold of total energy was set to 0.6 eV, 0.2 eV, 0.3 eV and 0.2 eV.

Following is an example input for SHARC dynamics:
\begin{verbatim}
printlevel 2
geomfile "geom"
veloc external
velocfile "veloc"

nstates 3 0 0 
actstates 3 0 0 
state 2 mch
coeff auto
rngseed -28624

ezero   -536.9454713000
tmax 500.000000
stepsize 0.500000
nsubsteps 25

surf diagonal
coupling overlap
ekincorrect parallel_vel
reflect_frustrated none
decoherence_scheme edc
decoherence_param 0.1
hopping_procedure sharc
\end{verbatim}

\subsection{Complete Active Space Self-Consistent Field (CASSCF)}

In quantum chemistry, accurately capturing electron correlation---the interaction of electrons relative to one another---is pivotal for an in-depth understanding of a molecule's electronic structure. While standard methods like Hartree-Fock (HF) have their strengths, they can falter in specific scenarios. This is where the CASSCF method becomes instrumental.

Central to CASSCF is the categorization of molecular orbitals into three distinct groups:

\begin{enumerate}
    \item \textbf{Inactive (core) orbitals:} These are fully occupied orbitals, exempted from the correlation treatment.
    \item \textbf{Active orbitals:} A defined number of electrons within these orbitals undergo correlation across a predetermined set of orbitals. The flexibility in electron configuration within the active space encapsulates static electron correlation.
    \item \textbf{Virtual (secondary) orbitals:} Remaining unoccupied, these orbitals are sidelined from the primary correlation procedure.
\end{enumerate}

The CASSCF methodology initially optimizes the active space orbitals employing a comprehensive configuration interaction (CI) calculation. This act of considering all plausible electron configurations within the active ambit captures static correlation. To address dynamic correlation, supplementary methods, like multi-reference perturbation theory (MRPT), are often invoked.

\textit{Advantages of CASSCF:}
\begin{itemize}
    \item Offers a harmonized treatment of electron correlation.
    \item Particularly apt for systems with closely-spaced electronic states, encompassing transition states, metal complexes, and excited states.
\end{itemize}

However, one should note the substantial computational demands, especially with enlarging active spaces, which can potentially restrict its application or mandate approximate solutions.

\begin{figure}[ht]
\centering
\includegraphics[width=1\textwidth]{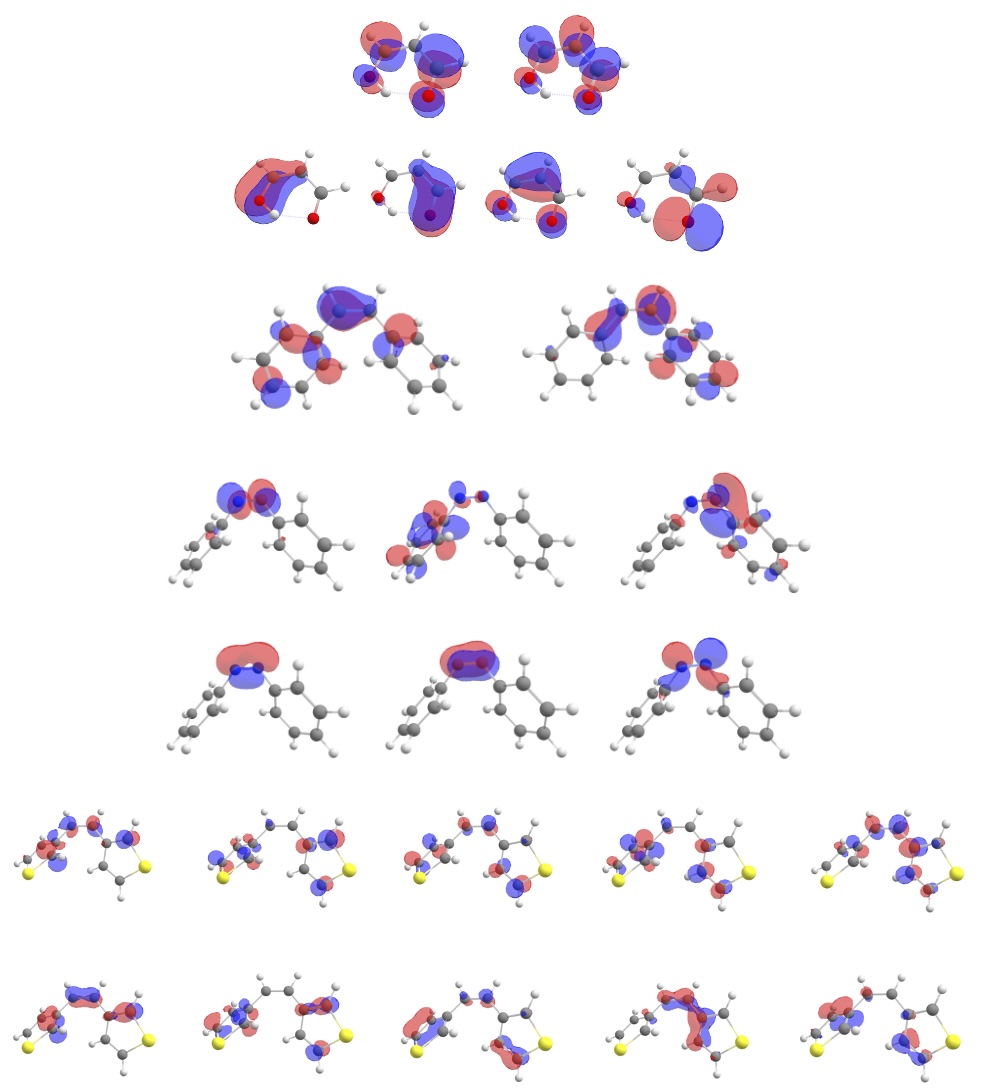}
\caption{Active space orbitals used SA-CASSCF calculations for malonaldehyde, stilbene, azobenzene and dithiophine.}
\label{fig:s_azo_as}
\end{figure}

For SA-CASSCF calculations, OpenMolcas 22.06 was used, which is available at \url{https://gitlab.com/Molcas/OpenMolcas}. The active space orbitals of the starting configurations are listed as following (Figure \ref{fig:s_azo_as}).

\subsection{Unrestricted KS-DFT}

In molecular modeling, the precise representation of electronic configurations during chemical reactions is paramount. The popular restricted KS-DFT inherently pairs electrons, enforcing identical spatial orbitals for both spin-up and spin-down states. 

Consider the paradigmatic dissociation of hydrogen (H$_2$) into atomic hydrogen:
\begin{equation}
\text{H}_2 \rightarrow 2\text{H}
\end{equation}

Within the confines of restricted KS-DFT, as H$_2$ dissociates, the emerging electrons---now localized on individual atoms---are still bound to identical spatial distributions. This treatment may distort the real physical scenario.

Unrestricted KS-DFT, on the other hand, permits differentiation between spin-up and spin-down spatial orbitals, enabling a nuanced portrayal of the process. In the H$_2$ example, would independently model the electron on each hydrogen atom, providing a truer representation of the physical system. 

For all unrestricted KS-DFT calculations, we used M06\cite{zhao2008m06} meta-GGA hybrid functional with 6-31g basis set. All calculations are done with the Psi4\cite{turney2012psi4} package (available at \url{https://github.com/psi4/psi4}) interfaced the ASE\cite{larsen2017atomic} package (available at \url{https://github.com/rosswhitfield/ase}).

\subsection{Dihydrogen dissociation: a comparative case of RKS, UKS and CASSCF}

\begin{figure}[ht]
\centering
\includegraphics[width=0.6\textwidth]{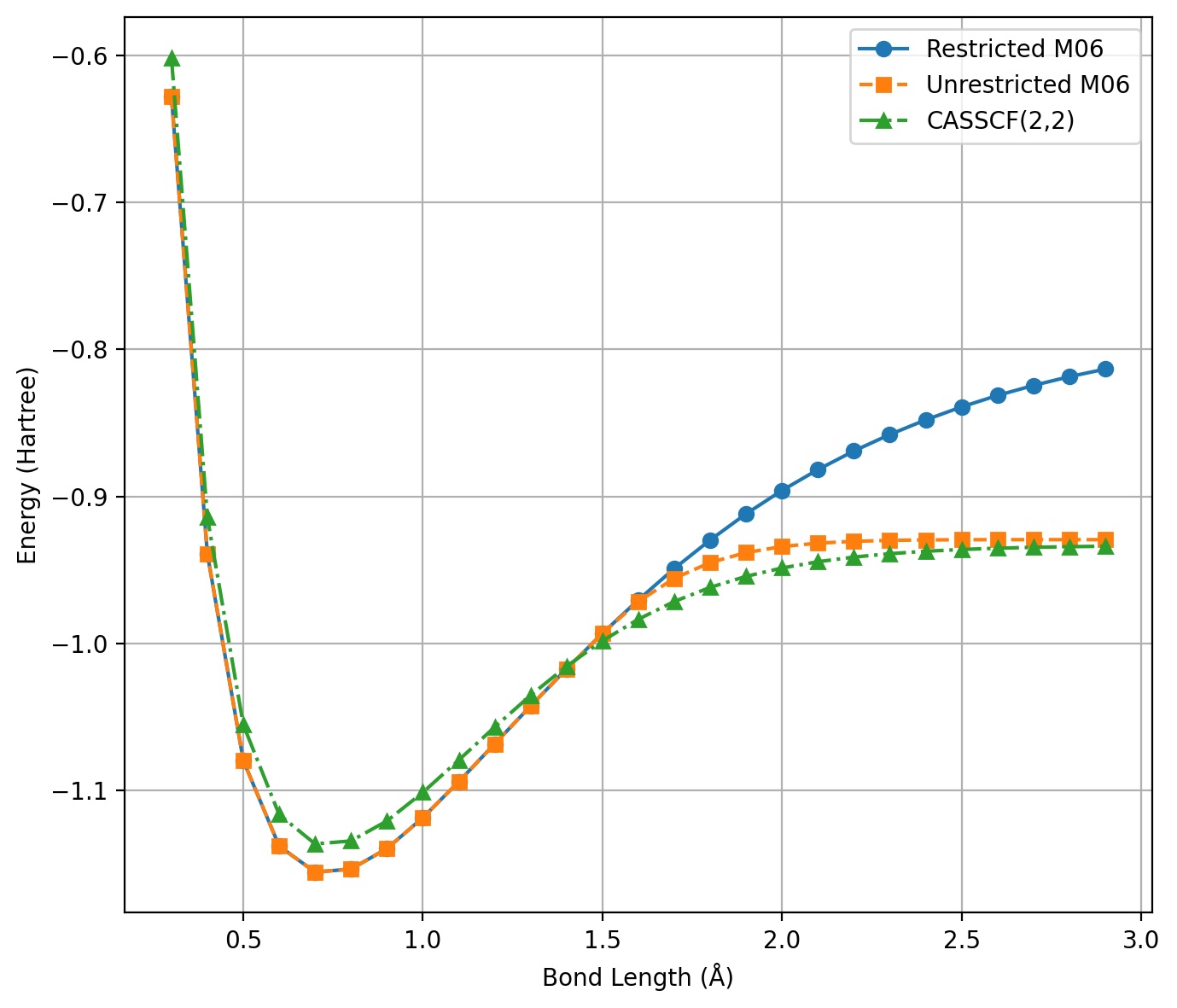}
\caption{Dissociation Curve of dihydrogen molecule using RKS, UKS, and CASSCF(2,2) methods. RKS is inherently inadequate for capturing the true electronic structure nuances of bond-breaking events, as seen in the deviation from the CASSCF. In principle, multi-reference methods are essential for accurate modeling of such chemical reactions, ensuring a more holistic representation of the electronic correlation effects.}
\label{fig:h2pes}
\end{figure}

The limitation of using DFT, espeicially restricted DFT becomes evident when examining the H-H bond-breaking process, as illustrated in Figure \ref{fig:h2pes}. Here, spin-unpolarized DFT yields an inaccurate yet smooth curve when juxtaposed against its spin-polarized counterpart, with CASSCF serving as the reference.

\bibliography{sample}